\documentclass{llncs}

\usepackage{graphicx}
\usepackage{tikz}
\usepackage{comment}
\usepackage{amsmath,amssymb}
\usepackage{color}
\usepackage[accsupp]{axessibility}
\usepackage{enumitem}
\usepackage{diagbox}
\usepackage{multirow}
\usepackage{colortbl}
\usepackage{booktabs}
\usepackage{xspace}
\usepackage{xcolor}
\usepackage{tcolorbox}
\usepackage{soul}
\usepackage{hyperref}
\usepackage{microtype} 
\usepackage[noabbrev]{cleveref} 

\newlist{todolist}{itemize}{2}
\setlist[todolist]{label=$\square$}

\newenvironment{todo}{%
    \begin{tcolorbox}[colback=red!5!white,colframe=red!75!black,title=TODO]
    \begin{todolist}
}{%
    \end{todolist}
    \end{tcolorbox}
}

\newcommand{\fixme}[1]{\xspace\textcolor{red}{\hl{\textbf{#1}}}}

\newcommand{\nilaksh}[1]{\fixme{#1 -ND}}

\newcommand{\MKCLEAN}{
    \renewcommand{\fixme}[1]{}
    
}
\MKCLEAN 

\definecolor{mtlorange}{rgb}{1,0.918,0.773}
\newcommand{\hlo}[1]{{\sethlcolor{mtlorange}\hl{#1}}}

\def\ybarz{\bar{y}_{z}}
\def\ybarx{\bar{y}_{x}}
\def\yhatx{\hat{y}_{x}}
\def\kbarz{\bar{k}_{z}}

\newcommand{\etal}{\textit{et al}.}
\newcommand{\ie}{\textit{i}.\textit{e}., }

\begin{document}
\pagestyle{plain}
\mainmatter

\title{SkeleVision: Towards Adversarial Resiliency of Person Tracking with Multi-Task Learning}
\titlerunning{}

\author{Nilaksh Das \and
Sheng-Yun Peng \and
Duen Horng Chau}
\authorrunning{}
\institute{Georgia Institute of Technology\\
\email{\{nilakshdas,speng65,polo\}@gatech.edu}}

\maketitle

\begin{abstract}
Person tracking using computer vision techniques
has wide ranging applications such as
autonomous driving, home security and sports analytics.
However, the growing threat of adversarial attacks
raises serious concerns regarding the security and reliability
of such techniques.
In this work, we study the impact of
multi-task learning (MTL) on the adversarial robustness
of the widely used SiamRPN tracker,
in the context of person tracking.
Specifically, we investigate the effect of
jointly learning
with semantically analogous tasks
of person tracking and human keypoint detection.
We conduct extensive experiments
with more powerful adversarial attacks
that can be physically realizable,
demonstrating the practical value of our approach.
Our empirical study
with simulated
as well as real-world datasets
reveals that 
training with MTL
consistently makes it harder to attack
the SiamRPN tracker,
compared to typically training
only on the single task of person tracking.

\end{abstract}

\section{Introduction}

Person tracking is extensively used in various
real-world use cases such as 
autonomous driving~\cite{bhattacharyya2018long,yagi2018future,yao2019unsupervised},
intelligent video surveillance~\cite{bohush2019robust,ye2020person,ahmed2021real}
and sports analytics~\cite{bridgeman2019multi,kong2020long,liang2020multi}.
However, vulnerabilities in the underlying techniques 
revealed by a growing body of adversarial ML research
\cite{szegedy2013intriguing,goodfellow2014explaining,madry2017towards,chen2018shapeshifter,eykholt2018robust,wiyatno2019physical,jia2020fooling,xu2020adversarial,chen2021unified}
seriously calls into question
the trustworthiness of these techniques
in critical use cases.
While several methods
have been proposed to
mitigate threats from adversarial attacks in general
\cite{cisse2017parseval,madry2017towards,tramer2017ensemble,schmidt2018adversarially,simon2019first},
defense research in the tracking domain
remains sparse~\cite{jia2020robust}.
This is especially true for the new generation
of physically realizable attacks~\cite{chen2018shapeshifter,eykholt2018robust,wiyatno2019physical}
that pose a greater threat to real-world applications.

\begin{figure}[t]
    \centering
    \includegraphics[width=0.78\textwidth]{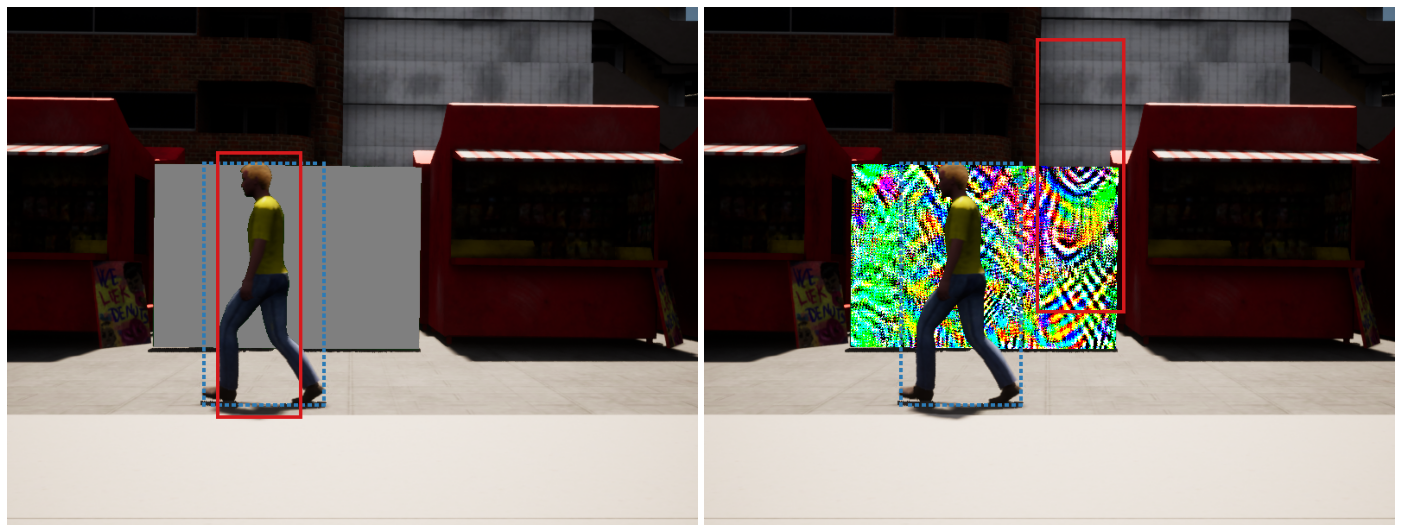}
    \caption{Example of a physically realizable patch attack. The dashed blue box shows the ground-truth bounding box and the solid red box shows the bounding box predicted by SiamRPN. In the benign case (left), the tracker is able to correctly track the person whereas in the adversarial case (right) the tracker is fooled by the adversarial patch.}
    \label{fig:physical_attack}
    \vspace{-0.4em}
\end{figure}

In this work, we aim to investigate the robustness characteristics
of the SiamRPN tracker~\cite{li2018high},
which is widely used in the tracking community.\nilaksh{add citations}
Specifically, our goal is to improve the tracking robustness
to a physically realizable patch attack~\cite{wiyatno2019physical}.
Such attacks are unbounded in the perceptual space
and can be deployed in realistic scenarios,
making them more harmful than imperceptible digital perturbation attacks.
\Cref{fig:physical_attack} shows an example
of such a physically realizable patch attack
that blends in the background.

\textbf{Multi-task learning (MTL)} has recently been touted
to improve adversarial robustness
to imperceptible digital perturbation attacks
for certain computer vision tasks~\cite{mao2020multitask,ghamizi2021adversarial}.
However, it is unclear if these proposed methods
translate to physically realizable attacks.
Moreover, these methods have primarily been studied
in the context of a single backbone branch
with one-shot inference,
whereas the Siamese architecture of the SiamRPN tracker
involves multiple branched stages,
posing interesting design considerations.
In this work, we aim to address these research gaps
by focusing on improving single-person tracking robustness.

As physically realizable attacks are unbounded in the perceptual space,
they can create easily perceptible, but inconspicuous perturbations
that fools a deep neural network into making incorrect predictions.
However humans can ignore such perturbations
by processing semantic knowledge of the real world.
This calls for implicitly incorporating some inductive biases 
that supervise the neural network to learn semantic constraints 
that humans so instinctively interpret.
To this effect, 
in this work we study the impact of MTL
on robustness of person tracking
with a semantically analogous task
such as human keypoint detection.

\smallskip
\noindent \textbf{Contributions}
\begin{itemize}[label={$\bullet$},leftmargin=*,topsep=0pt]
\item \textbf{First Study of Tracking Robustness with MTL.}
To the best of our knowledge,
our work is the first to uncover
the robustness gains
from MTL
in the context of person tracking
for physically realizable attacks.
Our code is made available at
\url{https://github.com/nilakshdas/SkeleVision}.

\item \textbf{Novel MTL Formulation for Tracking.}
We augment the SiamRPN tracker for MTL 
by attaching a keypoint detection head
to the template branch of the shared backbone
while jointly training.

\item \textbf{Extensive Evaluation.}
We conduct extensive experiments 
to empirically evaluate the effectiveness of our MTL approach   
by varying attack parameters,
network architecture, and MTL hyperparameters.

\item \textbf{Discovery.}
Our experiments
with simulated
and real-world datasets
reveal that 
training with MTL
consistently makes it harder to attack
the SiamRPN tracker
as compared to training
only on the single task of person tracking.
\end{itemize}

\section{Related Work}

Since its inception with SiamFC~\cite{bertinetto2016fully}, 
the Siamese architecture has been 
leveraged by multiple real-time object trackers 
including DSiam~\cite{guo2017learning}, SiamRPN~\cite{li2018high}, DaSiamRPN~\cite{zhu2018distractor}, SiamRPN++~\cite{li2019siamrpn++}, SiamAttn~\cite{yu2020deformable} 
and SiamMOT~\cite{shuai2021siammot}. 
In this work, we experiment with SiamRPN as the target tracker 
since many other trackers share a similar network architecture as SiamRPN, 
and the properties of SiamRPN can be generalized to other such state-of-the-art trackers. 

\subsection{Multi-task Learning}
MTL aims 
to learn multiple related tasks 
jointly to improve 
the generalization performance 
of all the tasks~\cite{zhang2021survey}. 
It has been applied 
to various computer vision tasks
including image classification~\cite{luo2013manifold},
image segmentation~\cite{mao2020multitask},
depth estimation~\cite{liebel2019multidepth},
and human keypoint detection~\cite{kocabas2018multiposenet}.

MTL has also been introduced 
for the video object tracking task~\cite{kristan2019seventh,kristan2020eighth,kristan2021ninth}. 
Zhang \etal~\cite{zhang2013robust,zhang2017multi,zhang2018learning}
formulate the particle filter tracking as a structured MTL problem, 
where learning the representation of each particle 
is treated as as an individual task. 
Wang \etal~\cite{wang2021towards} show 
that joint training of natural language processing 
and object tracking can 
link the local and global search together, 
and lead to a better tracking accuracy.
Multi-modal RGB-depth and RGB-infrared tracking 
also demonstrate that
including the depth or infrared information 
in the tracking training process 
can improve the overall performances~\cite{yan2021depthtrack,zhu2022visual,zhang2019siamft,zhang2020dsiammft}.

\subsection{Adversarial Attacks}
Machine learning model are easily fooled by adversarial attacks~\cite{dong2018boosting}. 
Adversarial attacks can be classified as
digital perturbation attacks~\cite{szegedy2013intriguing,goodfellow2014explaining,madry2017towards} 
and physically realizable attacks~\cite{chen2018shapeshifter,eykholt2018robust,wiyatno2019physical,xu2020adversarial}. 
In the tracking community, multiple attacks have been proposed 
to fool the object tracker~\cite{jia2020fooling,chen2021unified}.
Fast attack network~\cite{liang2020efficient} 
attacks the Siamese network based trackers 
using a drift loss and embedded features. 
The attack proposed by Jia \etal~\cite{jia2021iou}
degrades the tracking accuracy through an IoU attack, 
which sequentially generates perturbations 
based on the predicted IoU scores. 
The attack requires ground-truth when performing the attack.
Wiyatno and Xu~\cite{wiyatno2019physical} propose a method 
to generate an adversarial texture. 
The texture can lock the GOTURN tracker~\cite{held2016learning} 
when a tracking target moves in front of it.

\subsection{Adversarial Defenses in Tracking}
General defense methods for computer vision tasks 
include adversarial training~\cite{tramer2017ensemble}, 
increasing labeled and unlabeled training data~\cite{schmidt2018adversarially}, 
decreasing the input dimensionality~\cite{simon2019first}, 
and robust optimization procedures~\cite{yan2018deep,xu2020adversarial_attack}. 
However, not many defense methods have been proposed 
to improve the tracking robustness under attack. 
Jia \etal~\cite{jia2020robust} attempt to eliminate 
the effect of the adversarial perturbations 
via learning the patterns from the attacked images.
Recently, MTL has been shown to improve 
the overall network robustness~\cite{ghamizi2021adversarial}, 
especially in image segmentation~\cite{mao2020multitask} 
and text classification~\cite{liu2017adversarial}. 
Our work is the first that studies MTL for person tracking
with a physically realizable attack. 
\section{Preliminaries}

The input to the tracker
can be denoted as $\{x, z, \ybarx, \ybarz\}$,
where $x$ is the detection frame in which the subject is to be tracked, 
$z$ is the template frame containing an exemplar representation of the subject,
and respectively, $\ybarx$ and $\ybarz$ are the ground-truth bounding box coordinates
within the corresponding frames.

\subsection{Tracking with SiamRPN}
\label{sec:siamrpn}

\begin{figure}[t]
    \centering
    \includegraphics[trim=0cm 0.8cm 0cm 0cm, width=\textwidth]{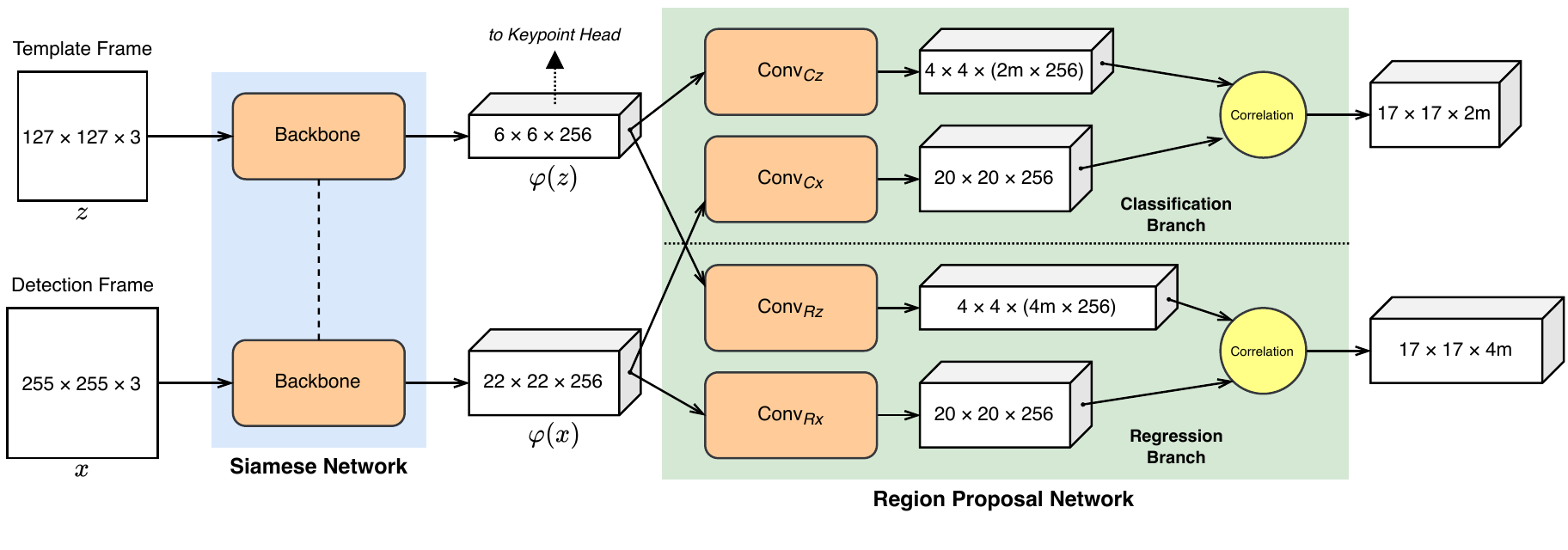}
    \caption{
    Overview of the SiamRPN architecture for tracking. 
    For multi-task learning, the output of the template branch
    is passed to a keypoint head for keypoint detection.
    }
    \label{fig:siamrpn_arch}
\end{figure}

In this work, we focus on the Siamese-RPN model (SiamRPN)~\cite{li2018high},
which is a widely used tracking framework
based on the Siamese architecture\nilaksh{needs citation}.
An overview of the SiamRPN architecture is shown in \Cref{fig:siamrpn_arch}.
SiamRPN consists of 
a Siamese network for extracting features
and a region proposal network (RPN), 
also referred to as the RPN head,
for predicting bounding boxes.

The Siamese network has two branches:
(1) the \textit{template branch} 
which receives a template patch $z' = \Gamma(z, \ybarz, s_z)$ as input;
and (2) the \textit{detection branch}
which receives a detection patch $x' = \Gamma(x, \ybarx, s_x)$ as input.
Here, $\Gamma(\cdot)$ is simply a crop operation
that ensures only a limited context of size $s$
centered on the bounding box $y$ 
is passed to the network~\cite{li2018high}.
The corresponding sizes $s_z$ and $s_x$
are shown in \Cref{fig:siamrpn_arch}.
For notational convenience,
we use $z$ for $z'$ and $x$ for $x'$
hereon.
The two branches of the Siamese network
use a shared backbone model
such that inputs to both branches
undergo the same transformation $\varphi(\cdot)$.
Hence, we can denote the output feature maps of the Siamese network
as $\varphi(z)$ and $\varphi(x)$ 
for the template and detection branches, respectively.
In this work, we use the SiamRPN model
with AlexNet backbone~\cite{krizhevsky2012imagenet}.

The RPN head can also be separated into two branches 
as shown in \Cref{fig:siamrpn_arch}.
Considering $m$ anchors distributed across the detection frame,
the classification branch predicts whether 
each respective anchor is a background or foreground anchor. 
Hence, the classification branch has $2m$ output channels
corresponding to $m$ anchors.
The regression branch on the other hand
predicts 4 box coordinate regression deltas~\cite{ren2015faster}
for each anchor,
and therefore has $4m$ output channels.

While training, the classification and regression branches of the RPN head
yield $\mathcal{L}_{cls}$ and $\mathcal{L}_{reg}$ respectively,
where $\mathcal{L}_{cls}$ is the cross-entropy loss 
and $\mathcal{L}_{reg}$ is a smooth $L_1$ loss~\cite{li2018high}.
Finally, the total weighted loss optimized for is as follows:
\begin{align}
    \mathcal{L}_{TRK}(x, z, \ybarx) = \lambda_C \mathcal{L}_{cls}(x, z, \ybarx) + \lambda_R \mathcal{L}_{reg}(x, z, \ybarx)
\end{align}

During inference, the network acts as a single-shot detector.
Typically, a sequence of frames $\textbf{x} = \{x_1, \dots x_n\}$ is provided with 
the ground-truth bounding box coordinates $\bar{y}_{x_1}$ of the first frame as input.
Hence, the first frame $x_1$  becomes the template frame $z$
used to compute the feature map $\varphi(z)$ once,
which can be considered as detector parameters
for predicting bounding box coordinates 
for input frames from the same sequence.
We denote the predicted bounding box 
for an input frame as $\yhatx$.
As mentioned previously, SiamRPN crops the context 
centered on the ground-truth bounding box.
For inference, the context region 
is determined by the predicted bounding box of the previous frame.
Finally, the tracking performance is evaluated
using a mean intersection-over-union (mIoU) metric
of the predicted and ground-truth bounding boxes
across all frames from all input sequences.

\subsection{Multi-task Learning with Shared Backbone}
\label{sec:keypoint_head}

To provide semantic regularization for tracking,
we perform joint multi-task training
by attaching a fully convolutional keypoint prediction head
to the template branch of SiamRPN.
Our hypothesis is that joint training
with an additional task head
attached to the shared backbone
would encourage the backbone to 
learn more robust features~\cite{ilyas2019adversarial}
for facilitating multiple tasks.
Since the shared backbone is also used 
during tracking inference,
the learned robust features can 
make it harder for adversarial perturbations
to fool the model.
We select the task of human keypoint prediction for this purpose
as it is more semantically analogous to the task of person tracking.

The keypoint head is attached to the template branch
as it has a more focused context~\cite{li2018high}.
Therefore, the keypoint head receives 
$\varphi(z)$ as input.
The keypoint head network consists of convolutional blocks 
followed by a transpose convolution operation 
that ``upsamples'' the intermediate feature map 
to an expanded size
with number of output channels equaling
the number of keypoints being predicted.
Finally, bilinear interpolation is performed
to match the size of the input frame.
The resulting feature volume has a shape of
$H \times W \times K$,
where $H$ and $W$ are 
the height and width of 
the input frame respectively,
and $K$ is the number of keypoints.
Hence, each position in the $K$-channel dimension
corresponds to a keypoint logit score.
Given the ground-truth keypoints $\kbarz$,
the binary cross-entropy loss is computed
with respect to each position in the channel dimension.
We denote this as $\mathcal{L}_{KPT}$.
For multi-task training, the total loss is a weighted sum:
\begin{align}
    \mathcal{L}_{MTL}(x, z, \ybarx, \kbarz) = \mathcal{L}_{TRK}(x, z, \ybarx) + \lambda_K \mathcal{L}_{KPT}(z, \kbarz)
\end{align}

The ground-truth keypoint annotation also consists of a visibility flag
that allows us to suppress spurious loss from being backpropagated
for keypoints that are occluded or not annotated.

\subsection{Adversarial Attacks}

Adversarial attacks introduce malicious perturbations
to the input samples in order to
confuse the tracker into making incorrect predictions.
In this work, we use white-box untargeted attacks
that aim to reduce the tracking performance
by minimizing the mIoU metric.
Adversarial attacks target a task loss,
whereby the objective is to increase the loss
by performing gradient ascent\nilaksh{add citations}.
Given the predicted and ground-truth bounding boxes
$\yhatx$ and $\ybarx$ respectively,
we use the L1-loss as the task loss as proposed in ~\cite{wiyatno2019physical}
for attacking an object tracker:

\begin{align}
\mathcal{L}_{ADV}(\yhatx, \ybarx) = \|\yhatx - \ybarx\|_1 \label{eq:task_loss}
\end{align}

Based on means of application of the adversarial perturbation 
and additional constraints placed on the perturbation strength,
attacks can be further classified into two distinct types:

\smallskip

\noindent
\textbf{Digital Perturbation Attacks.}
These attacks introduce fine-grained pixel perturbations
that are imperceptible to humans
\cite{szegedy2013intriguing,goodfellow2014explaining,madry2017towards}.
Digital perturbation attacks can manipulate any pixel of the input,
but place imperceptibility constraints 
such that the adversarial output $x_{adv}$
is within an $l_p$-ball of the benign input $x_{ben}$,
\ie $\|x_{adv} - x_{ben}\|_p \leq \epsilon$.
Such attacks, although having high efficacy,
are considered to be physically non-realizable.
This is due to 
the spatially unbounded granular pixel manipulation
of the attack
as well as the fact
that a different perturbation
is typically applied to each frame
of a video sequence.

\smallskip
\noindent
\textbf{Physically Realizable Attacks.}
These attacks place constraints 
on the input space that can be manipulated
by the attack
\cite{chen2018shapeshifter,eykholt2018robust,wiyatno2019physical,xu2020adversarial}.
In doing so, the adversarial perturbations
can be contained within realistic objects
in the physical world, 
such as a printed traffic sign~\cite{chen2018shapeshifter} 
or a T-shirt~\cite{xu2020adversarial}.
As an attacker can completely control
the form of the physical adversarial artifact,
physically realizable attacks are unbounded
in the perceptual space and place no constraints
on the perturbation strength.
In this work, we consider
a physically realizable attack
based on~\cite{wiyatno2019physical}
that produces a background patch perturbation
to fool an object tracker (\Cref{fig:physical_attack}).
It is an iterative attack that
follows gradient ascent
for the task loss described in~\Cref{eq:task_loss}
by adding a perturbation to the input
that is a product of the input gradient  
and a step size $\delta$:

\begin{align}
    x^{(i)} = x^{(i-1)} + \delta \nabla_{x^{(i-1)}} \mathcal{L}_{ADV} \label{eq:att_iter}
\end{align}

\section{Experiment Setup}

We perform extensive experiments
and demonstrate that models 
trained with MTL
are more resilient to 
adversarial attacks.
The multi-task setting consists of
jointly training a shared backbone
for semantically related tasks
such as person tracking
and human keypoint detection
(\Cref{sec:exp_architecture}).\nilaksh{add description of other subsections}
We evaluate the tracking robustness on
a state-of-the-art physically realizable adversarial attack
for object trackers.
We test our models on a photo-realistic simulated dataset
as well as a real-world video dataset for person tracking
(\Cref{sec:exp_evaluation}).

\subsection{Architecture}
\label{sec:exp_architecture}

For tracking,
we leverage a SiamRPN model
(\Cref{fig:siamrpn_arch})
with an AlexNet backbone
and an RPN head
as described in~\cite{li2018high}.
The inputs to the model
are a template frame ($127 \times 127$)
and a detection frame ($255 \times 255$),
fed to the backbone network.
Finally, the RPN head of the
model produces classification
and localization artifacts
corresponding to $m=5$ anchors
for each spatial position.
The anchors have aspect ratios of
$\{0.33, 0.5, 1, 2, 3\}$ respectively.

A keypoint head
is also attached to the template branch
of the network,
\ie the keypoint head receives
the activation map with dimensions 
$6 \times 6 \times 256$
as input.
We attach the keypoint head
to the template branch
as the template frame
has a more focused context,
and typically has only one subject
in the frame,
leading to more stable keypoint training.
The base keypoint head has
2 convolutional blocks
with $\{128, 64\}$ channels respectively.
We also perform ablation experiments
by increasing the depth of the keypoint head
to 4 blocks with
$\{128, 128, 64, 64\}$ channels respectively
(\Cref{sec:res_keypoint_head_depth}).
The convolutional blocks are followed by a
transpose convolution block
with 17 output channels,
which is the same as
the number of human keypoints
represented in the MS COCO format~\cite{lin2014coco}.
Bilinear interpolation is performed
on the output of the transpose convolution block
to expand the spatial output dimensions,
yielding an output with dimensions 
$127 \times 127 \times 17$.
Hence each of the 17 channels
correspond to spatial logit scores for the 17 keypoints.
\subsection{Training Data}
\label{sec:exp_train_data}

\begin{figure}[t]
    \centering
    \begin{minipage}[t][1\width]{0.2\textwidth}
        \centering
        \includegraphics[width=\textwidth]{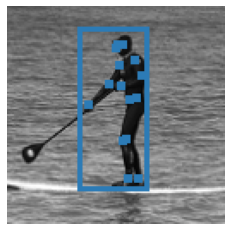}
    \end{minipage}
    \begin{minipage}[t][1\width]{0.2\textwidth}
        \centering
        \includegraphics[width=\textwidth]{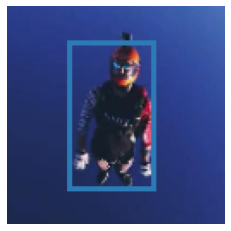}
    \end{minipage}
    \caption{Annotated training examples from the MS COCO (left) and LaSOT (right) datasets for person tracking. MS COCO has additional human keypoint annotations.}
    \label{fig:train_anno}
\end{figure}

We found that there is a dearth of
publicly available tracking datasets
that support ad-hoc tasks 
for enabling multi-task learning.
Hence, for our MTL training,
we create a hybrid dataset
that enables jointly training
with person tracking
and human keypoint detection.
For human keypoint annotations,
we leverage the MS COCO dataset~\cite{lin2014coco}
which contains more than 
$200k$ images and $250k$ person instances,
each labeled with 17 human keypoints.
The MS COCO dataset also annotates
person bounding boxes that we use for the tracking scenario.
As the MS COCO dataset consists of single images,
there is no notion of temporal sequences in the input.
Hence, for person tracking, we leverage data augmentation
to differentiate the template and detection frames
for the person instance annotation from the same image.
Therefore, the MS COCO dataset
allows us to train both the RPN head and keypoint head
jointly for person tracking and human keypoint detection.
We use the defined train and val splits
for training and validation.
Additionally, we merge this data
with the Large-scale Single Object Tracking (LaSOT)
dataset~\cite{fan2019lasot}.
Specifically, we extract all videos
for the ``\textit{person}'' class
for training the person tracking network.
This gives us 20 video sequences,
of which we use the first 800 frames
from each sequence for training
and the subsequent 100 frames for the validation set.
Hence, the combined hybrid dataset 
from MS COCO and LaSOT
enables our multi-task training.
\Cref{fig:train_anno} shows 2 example frames
from MS COCO and LaSOT datasets.

\subsection{Multi-Task Training}
\label{sec:exp_training}

For the multi-task training,
we fine-tune a 
generally pre-trained SiamRPN object tracker
jointly for the tasks of 
person tracking and human keypoint detection.
As we are specifically interested
in the impact of multi-task training,
we use the same loss weights 
$\lambda_C$ and $\lambda_R$
as proposed in~\cite{li2018high}
for the tracking loss $\mathcal{L}_{TRK}$.
We perform an extensive sweep of the
MTL loss weight $\lambda_K$
associated with the keypoint loss $\mathcal{L}_{KPT}$
(\Cref{sec:res_mtl_weight}).
For the baseline, we perform single-task learning (STL)
for person tracking by dropping the keypoint head
and only fine-tuning the RPN head with the backbone,
\ie the STL baseline has $\lambda_K = 0$.
All STL and MTL models are trained 
with a learning rate of
$8\times10^{-4}$
that yields the best baseline tracking results
as verified using a separate validation set.
We also study the impact of pre-training
the keypoint head separately 
before performing MTL 
(\Cref{sec:res_keypoint_head_pretrain}).
For pre-training the keypoint head,
we drop the RPN head 
and freeze the parameters
of the backbone network.
This ensures that the RPN head parameters
are still compatible with the backbone
after pre-training the keypoint head.
The keypoint head
is pre-trained
with a learning rate of
$10^{-3}$.
We train all models
for 50 epochs and select
the models with best
validation performance
over the epochs.

\subsection{Evaluation}
\label{sec:exp_evaluation}

\begin{figure}[t]
    \centering
    \includegraphics[width=0.9\textwidth]{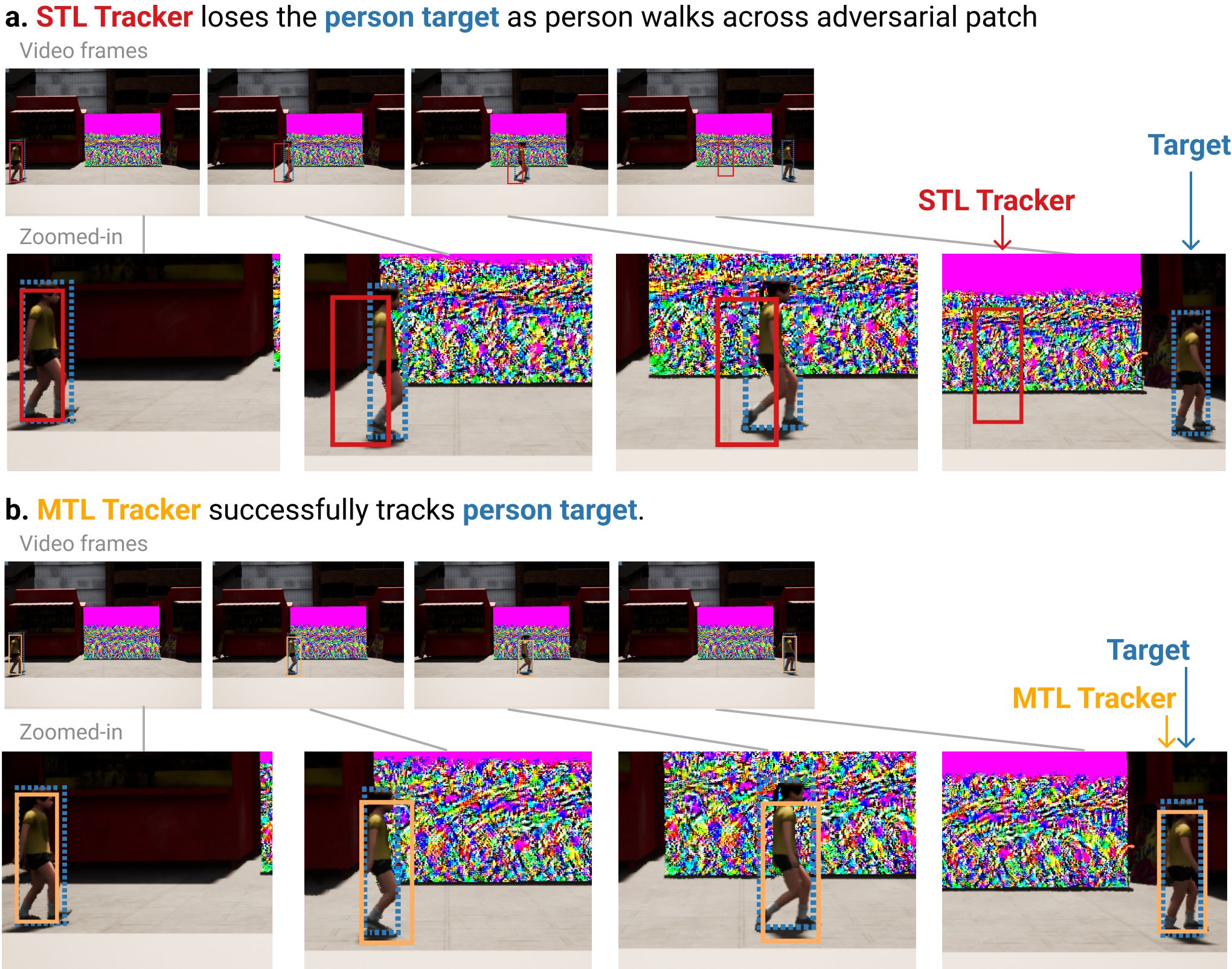}
    \caption{
    Example video frames from the ARMORY-CARLA dataset showing static adversarial patches for (a) STL and (b) MTL for an attack with $\delta=0.1$ and 10 steps. The patch is able to lock onto the STL tracker prediction (top), whereas the MTL tracker is consistently able to track the target (bottom). 
    }
    \label{fig:carla_attack}
\end{figure}

\begin{figure}[t]
    \centering
    \includegraphics[width=0.9\textwidth]{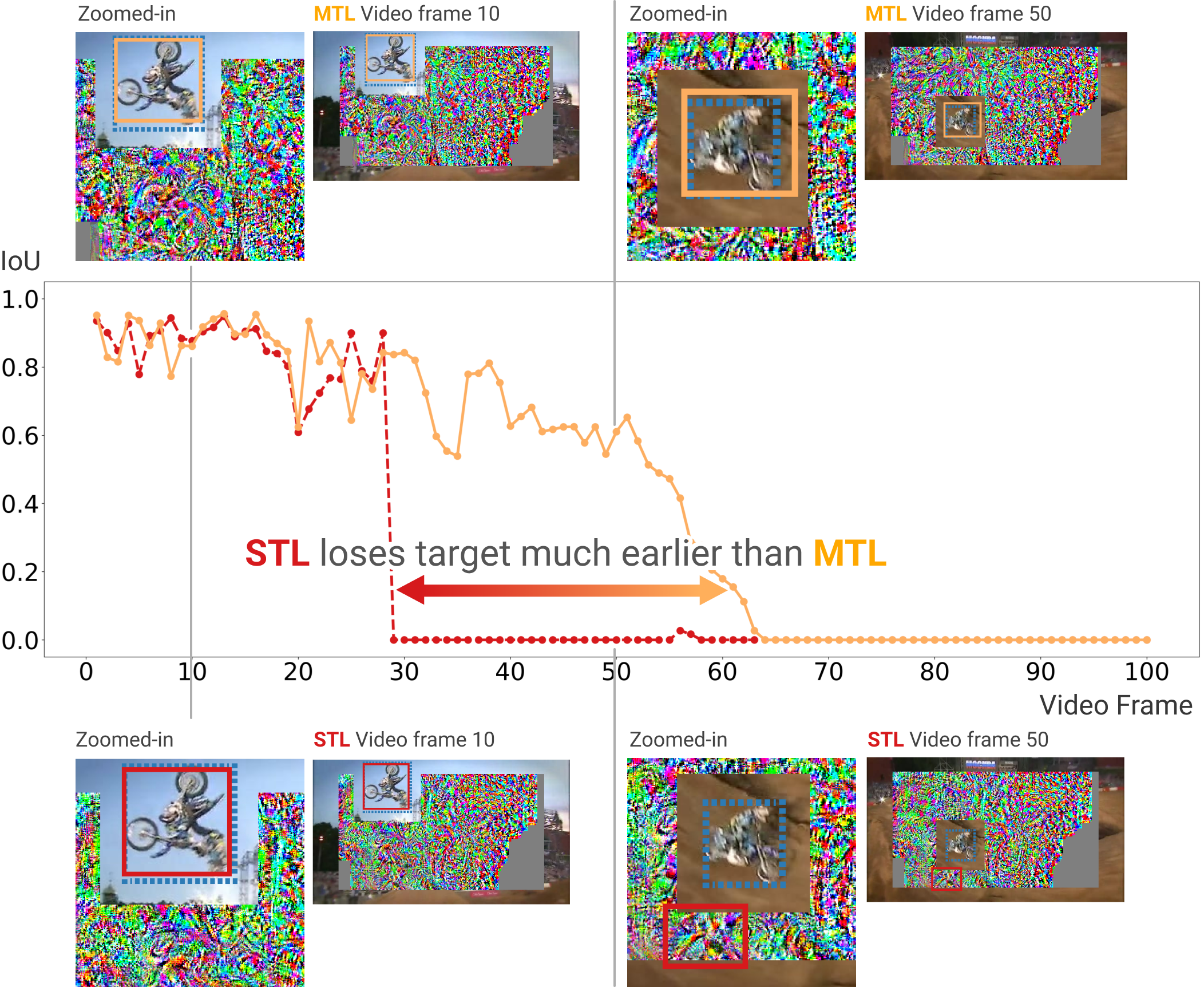}
    \caption{
    Example video frames and the corresponding adversarial IoU charts for the video from the OTB2015-Person dataset showing the constructed static adversarial patches for STL (red) and MTL (orange) for an attack with $\delta=0.1$ and 10 steps. The dashed blue box shows the ground-truth target. The attack misleads the STL tracker early, but struggles to mislead the MTL tracker until much later. The unperturbed gray regions in the patch are locations which are never predicted by the tracker. 
    }
    \label{fig:otb_attack}
\end{figure}

We evaluate our trained STL and MTL models
for the tracking scenario using the
mIoU metric
between ground-truth and predicted bounding boxes,
which is first averaged over all frames for a sequence,
and finally averaged over all sequences.

For testing the adversarial robustness 
of person tracking in a practical scenario,
we leverage a state-of-the-art 
physically realizable adversarial attack
for object trackers~\cite{wiyatno2019physical}.
The attack adds a static adversarial background patch
to a given video sequence that targets
the tracking task loss $\mathcal{L}_{ADV}$.
At each iteration of the attack,
gradient ascent is performed on
the task loss as per \Cref{eq:att_iter}
with a step size $\delta$.
In order to observe 
the effect of varying the step size
and attack iterations,
we experiment with multiple values 
and report results for
$\delta = \{0.1, 0.2\}$,
which we found to 
have stronger adversarial effect
on the tracking performance.
The attack proposed in ~\cite{wiyatno2019physical} 
has no imperceptibility constraints
and is unbounded in the perceptual space,
and can thus be considered 
an extremely effective adversarial attack.
As the attack relies on the gradients
of the task loss,
we implement an end-to-end 
differentiable inference pipeline
for the SiamRPN network
using the Adversarial Robustness Toolbox (ART) framework~\cite{art2018}.

\medskip

\noindent
We evaluate the adversarial robustness
of STL and MTL models
on 2 datasets:

\smallskip

\noindent
\textbf{ARMORY-CARLA.}
This is a simulated photo-realistic person tracking dataset
created using the CARLA simulator~\cite{dosovitskiy17carla},
provided by the ARMORY test bed~\cite{armory2022}
for adversarial ML research.
We use the ``\textit{dev}'' dataset split.
The dataset consists of 20 videos
of separate human sprites
walking across the scene
with various background locations.
Each video has an allocated patch
in the background 
that can be adversarially perturbed
to mimic a physically realizable attack
for person tracking.
The dataset also provides semantic segmentation annotations
to ensure that only the patch pixels in the background
are perturbed when a human sprite
passes in front of the patch.
\Cref{fig:carla_attack} shows
example video frames from the dataset
where this can be seen.
We find that the SiamRPN person tracker,
having been trained on real-world datasets,
has a reasonably high mIoU
for tracking the human sprites
when there is no attack performed;
thus qualifying the photo-realism 
of the simulated scenario.

\smallskip

\noindent
\textbf{OTB2015-Person.}
We use the Object Tracking Benchmark (OTB2015)~\cite{wu2015otb}
to test the robustness of MTL for person tracking
on a real-world dataset.
We extract all videos
that correspond to the task of person tracking,
which yields 38 videos
that we call the OTB2015-Person split.
As the dataset is intended for real-world tracking
and is not readily amenable to implement
physically realizable attacks,
we digitally modify the videos
for our attack to work.
For each video, we overlay a static background patch
that has a margin of 10\% from each edge,
covering 64\% of the total area
that can be perturbed by the attack.
Finally, for each frame of a video,
we only uncover the region annotated
by the ground-truth bounding box
with a padding of 25 pixels on each side.
Hence, the annotated subject 
is always completely visible
to the tracker with 
a digitally perturbed adversarial patch boundary.
\Cref{fig:otb_attack} shows example video frames
with the static patch attack as described here.
To ensure that the tracker
gets a clean ground-truth template,
we do not perturb the first frame.
Since this implements an unbounded digital attack
on inputs from the real-world perceptual space,
the attack is much stronger than
real-world physically realizable attacks.
For computational tractability,
we only attack the first 100 frames.

\section{Results}

\begin{figure}[t]
    \centering
    \includegraphics[width=\textwidth]{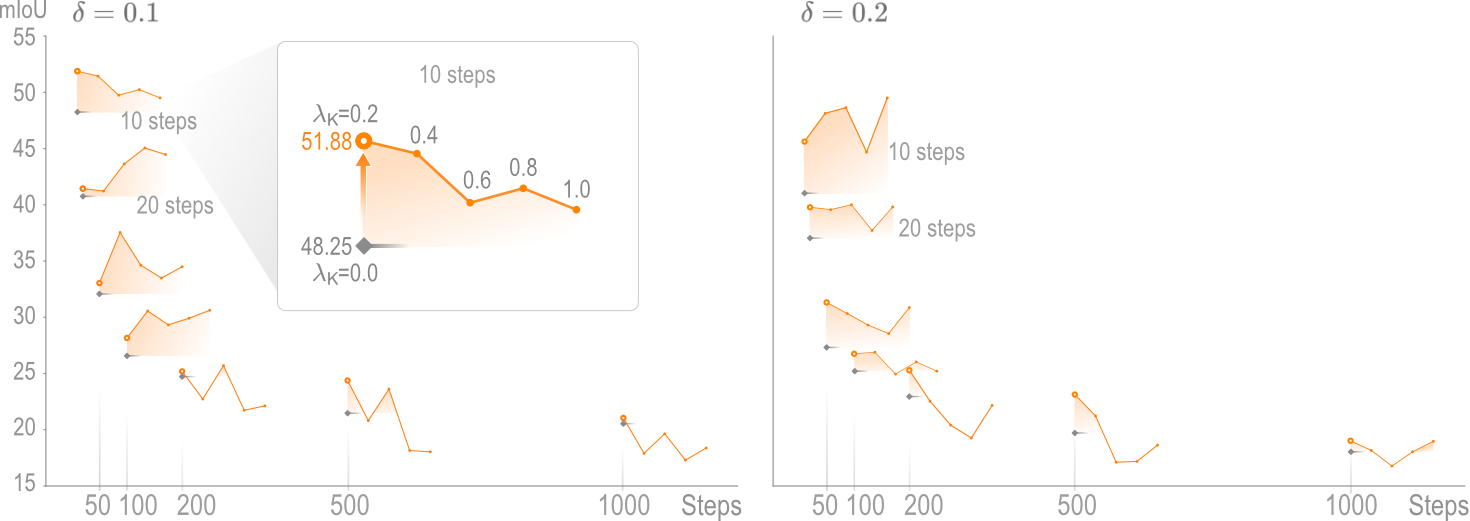}
    \caption{A unified visualization of the adversarial mIoU results
    from \Cref{tab:carla_patch} for the ARMORY-CARLA dataset
    with $\delta=0.1$ (left) and $\delta=0.2$ (right).
    The orange dots represent the MTL mIoU's and the gray flat lines
    represent the STL baseline mIoU's. 
    We see that the hollow orange dots ($\lambda_K=0.2$)
    are consistently above the STL baseline.
    }
    \label{fig:carla_sparklines}
\end{figure}

\begin{table}[t]
\centering
\caption{Adversarial mIoU results for ARMORY-CARLA dataset ($\uparrow$ is better). Values highlighted in orange show the cases in which MTL is more robust than STL. MTL model with $\lambda_K=0.2$ is consistently harder to attack than the STL model, and most often has the best performance. This table is also visualized in \Cref{fig:carla_sparklines} for clarity.}
\begin{tabular}{lc|c|ccccc}
\toprule
& & STL & \multicolumn{5}{c}{MTL} \\
& Steps \ & \ $\lambda_K = 0.0$ \ & \ $\lambda_K = 0.2$ & \ $\lambda_K = 0.4$ & \ $\lambda_K = 0.6$ & \ $\lambda_K = 0.8$ & \ $\lambda_K = 1.0$ \\
\midrule
\textit{benign} & 0 & 69.45 & {\cellcolor{mtlorange}}{69.59} & {\cellcolor{mtlorange}}{69.46} & {\cellcolor{mtlorange}}{69.70} & {\cellcolor{mtlorange}}{72.20} & {\cellcolor{mtlorange}}{\textbf{72.08}} \\
\midrule
\multirow{7}{*}{$\delta=0.1$} & 10 & 48.25 & {\cellcolor{mtlorange}}{\textbf{51.88}} & {\cellcolor{mtlorange}}{51.44} & {\cellcolor{mtlorange}}{49.74} & {\cellcolor{mtlorange}}{50.24} & {\cellcolor{mtlorange}}{49.50} \\
& 20 & 40.70 & {\cellcolor{mtlorange}}{41.44} & {\cellcolor{mtlorange}}{41.22} & {\cellcolor{mtlorange}}{43.63} & {\cellcolor{mtlorange}}{\textbf{45.05}} & {\cellcolor{mtlorange}}{44.47} \\
& 50 & 32.07 & {\cellcolor{mtlorange}}{33.04} & {\cellcolor{mtlorange}}{\textbf{37.54}} & {\cellcolor{mtlorange}}{34.63} & {\cellcolor{mtlorange}}{33.49} & {\cellcolor{mtlorange}}{34.49} \\
& 100 & 26.57 & {\cellcolor{mtlorange}}{28.16} & {\cellcolor{mtlorange}}{30.56} & {\cellcolor{mtlorange}}{29.33} & {\cellcolor{mtlorange}}{29.91} & {\cellcolor{mtlorange}}{\textbf{30.62}} \\
& 200 & 24.72 & {\cellcolor{mtlorange}}{25.19} & {\color{gray} 22.73} & {\cellcolor{mtlorange}}{\textbf{25.70}} & {\color{gray} 21.73} & {\color{gray} 22.12} \\
& 500 & 21.47 & {\cellcolor{mtlorange}}{\textbf{24.38}} & {\color{gray} 20.81} & {\cellcolor{mtlorange}}{23.61} & {\color{gray} 18.15} & {\color{gray} 18.04} \\
& 1000 & 20.54 & {\cellcolor{mtlorange}}{\textbf{21.05}} & {\color{gray} 17.90} & {\color{gray} 19.64} & {\color{gray} 17.30} & {\color{gray} 18.37} \\
\midrule
\multirow{7}{*}{$\delta=0.2$} & 10 & 41.03 & {\cellcolor{mtlorange}}{45.62} & {\cellcolor{mtlorange}}{48.13} & {\cellcolor{mtlorange}}{48.63} & {\cellcolor{mtlorange}}{44.68} & {\cellcolor{mtlorange}}{\textbf{49.50}} \\
& 20 & 37.04 & {\cellcolor{mtlorange}}{39.78} & {\cellcolor{mtlorange}}{39.57} & {\cellcolor{mtlorange}}{\textbf{40.00}} & {\cellcolor{mtlorange}}{37.72} & {\cellcolor{mtlorange}}{39.81} \\
& 50 & 27.32 & {\cellcolor{mtlorange}}{\textbf{31.32}} & {\cellcolor{mtlorange}}{30.33} & {\cellcolor{mtlorange}}{29.31} & {\cellcolor{mtlorange}}{28.55} & {\cellcolor{mtlorange}}{30.86} \\
& 100 & 25.24 & {\cellcolor{mtlorange}}{26.76} & {\cellcolor{mtlorange}}{\textbf{26.89}} & {\color{gray} 24.95} & {\cellcolor{mtlorange}}{26.03} & {\color{gray} 25.21} \\
& 200 & 22.95 & {\cellcolor{mtlorange}}{\textbf{25.29}} & {\color{gray} 22.54} & {\color{gray} 20.41} & {\color{gray} 19.27} & {\color{gray} 22.16} \\
& 500 & 19.71 & {\cellcolor{mtlorange}}{\textbf{23.13}} & {\cellcolor{mtlorange}}{21.23} & {\color{gray} 17.11} & {\color{gray} 17.18} & {\color{gray} 18.63} \\
& 1000 & 18.04 & {\cellcolor{mtlorange}}{\textbf{19.02}} & {\cellcolor{mtlorange}}{18.16} & {\color{gray} 16.77} & {\color{gray} 18.04} & {\cellcolor{mtlorange}}{18.97} \\
\bottomrule
\multicolumn{8}{l}{\scriptsize \hlo{orange} = MTL $>$ STL; {\color{gray} gray} = MTL $\leq$ STL; \textbf{bold} = highest in row}
\end{tabular}
\label{tab:carla_patch}
\end{table}
\begin{table}[t]
\centering
\caption{Adversarial mIoU results for OTB2015-Person dataset ($\uparrow$ is better). In most cases, MTL models are harder to attack compared to STL model, with $\lambda_K=0.2$ being most robust to the attack across several attack steps and step sizes.}
\begin{tabular}{lc|c|ccccc}
\toprule
& & STL & \multicolumn{5}{c}{MTL} \\
& Steps \ & \ $\lambda_K = 0.0$ \ & \ $\lambda_K = 0.2$ & \ $\lambda_K = 0.4$ & \ $\lambda_K = 0.6$ & \ $\lambda_K = 0.8$ & \ $\lambda_K = 1.0$ \\
\midrule
\textit{benign} & 0 & 69.42 & {\color{gray} 68.62} & {\color{gray} 67.84} & {\color{gray} 67.89} & {\color{gray} 65.97} & {\color{gray} 68.50} \\ 
\midrule
\multirow{5}{*}{$\delta$ = 0.1} & 10 & 54.29 & {\cellcolor{mtlorange}}{57.29} & {\cellcolor{mtlorange}}{57.78} & {\cellcolor{mtlorange}}{\textbf{57.95}} & {\cellcolor{mtlorange}}{57.09} & {\cellcolor{mtlorange}}{56.24} \\
 & 20 & 52.62 & {\cellcolor{mtlorange}}{55.45} & {\cellcolor{mtlorange}}{\textbf{55.68}} & {\cellcolor{mtlorange}}{55.56} & {\cellcolor{mtlorange}}{53.01} & {\cellcolor{mtlorange}}{52.68} \\
 & 50 & 48.54 & {\cellcolor{mtlorange}}{\textbf{52.84}} & {\cellcolor{mtlorange}}{52.33} & {\cellcolor{mtlorange}}{50.54} & {\cellcolor{mtlorange}}{50.94} & {\cellcolor{mtlorange}}{50.67} \\
 & 100 & 44.92 & {\cellcolor{mtlorange}}{\textbf{52.45}} & {\cellcolor{mtlorange}}{48.54} & {\cellcolor{mtlorange}}{48.63} & {\cellcolor{mtlorange}}{48.77} & {\cellcolor{mtlorange}}{48.25} \\
 & 200 & 45.40 & {\cellcolor{mtlorange}}{47.73} & {\cellcolor{mtlorange}}{\textbf{49.18}} & {\cellcolor{mtlorange}}{47.26} & {\cellcolor{mtlorange}}{46.50} & {\cellcolor{mtlorange}}{47.34} \\ 
\midrule
\multirow{5}{*}{$\delta$ = 0.2} & 10 & 53.93 & {\cellcolor{mtlorange}}{\textbf{57.65}} & {\cellcolor{mtlorange}}{56.60} & {\cellcolor{mtlorange}}{55.70} & {\cellcolor{mtlorange}}{56.46} & {\cellcolor{mtlorange}}{54.34} \\
 & 20 & 53.57 & {\cellcolor{mtlorange}}{55.21} & {\cellcolor{mtlorange}}{\textbf{56.16}} & {\cellcolor{mtlorange}}{56.08} & {\cellcolor{mtlorange}}{54.46} & {\color{gray} 52.86} \\
 & 50 & 49.15 & {\cellcolor{mtlorange}}{\textbf{52.81}} & {\cellcolor{mtlorange}}{51.12} & {\cellcolor{mtlorange}}{49.48} & {\cellcolor{mtlorange}}{50.71} & {\cellcolor{mtlorange}}{49.65} \\
 & 100 & 47.27 & {\cellcolor{mtlorange}}{\textbf{52.72}} & {\cellcolor{mtlorange}}{49.74} & {\cellcolor{mtlorange}}{48.13} & {\cellcolor{mtlorange}}{49.87} & {\cellcolor{mtlorange}}{47.81} \\
 & 200 & 46.19 & {\cellcolor{mtlorange}}{\textbf{51.05}} & {\cellcolor{mtlorange}}{48.83} & {\cellcolor{mtlorange}}{47.24} & {\cellcolor{mtlorange}}{48.26} & {\cellcolor{mtlorange}}{47.04} \\
\bottomrule
\multicolumn{8}{l}{\scriptsize \hlo{orange} = MTL $>$ STL; {\color{gray} gray} = MTL $\leq$ STL; \textbf{bold} = highest in row}
\end{tabular}
\label{tab:otb_patch}
\end{table}

Our experiments reveal
that models trained with MTL
consistently make it harder
for an adversarial attack to succeed
by making the shared backbone network
learn robust MTL features.
Given an iterative attack,
higher number of iterations corresponds
to increased attack difficulty
and higher attacker cost.
We report the mIoU for increasing attack steps
from $\{$10, 20, 50, 100, 200, 500, 1000$\}$
for the ARMORY-CARLA dataset in \Cref{tab:carla_patch}.
We also visually summarize 
these gains from the MTL approach
for the ARMORY-CARLA dataset in 
\Cref{fig:carla_sparklines}.
For $\lambda_K=1.0$, 
which is the base MTL setting,
the MTL model improves upon the benign mIoU
from 69.45 to 72.08.
Additionally, 
the MTL model is more robust
than the STL baseline
up to 100 attack steps for $\delta=0.1$
and 50 attack steps for $\delta=0.2$.
This implies the attack cost is higher
for attacking an MTL model
compared to its STL counterpart.
The mIoU for increasing attack steps
for the OTB2015-Person dataset
are shown in \Cref{tab:otb_patch}.
We observe a degradation in the benign
MTL performance in this case,
which may partly be attributed to
the resolution mismatch
between the high resolution
training examples~\cite{lin2014coco,fan2019lasot}
from MS COCO and LaSOT datasets,
compared to lower resolution 
evaluation videos samples
from the OTB2015 dataset.
In the adversarial case,
the base MTL model
is more robust than STL baseline
for up to 200 steps
for $\delta=0.1$.
For $\delta=0.2$,
the base MTL model
fails to show robustness
for 20 steps,
and slightly better robustness
for other attack steps.
We see further improvements 
in the adversarial resiliency
for varying $\lambda_K$,
discussed in \Cref{sec:res_mtl_weight}.

\subsection{Varying MTL Weight}
\label{sec:res_mtl_weight}

We study the effect of varying
the MTL weight $\lambda_K$,
which controls the amount of 
keypoint loss $\mathcal{L}_{KPT}$
that is backpropagated.
We train separate models by enumerating
$\lambda_K = \{0.2, 0.4, 0.6, 0.8, 1.0\}$,
and perform adversarial patch attack on
each model for multiple adversarial settings.
The results for ARMORY-CARLA and OTB2015-Person
are shown in \Cref{tab:carla_patch} and \Cref{tab:otb_patch}
respectively.
%
%
We find that for the given shallow
keypoint head architecture ($\{128, 64\}$ channels),
a lower value of $\lambda_K$
is more optimal
under adversarial attack.
For both datasets,
the MTL model with $\lambda_K=0.2$ 
is consistently harder to attack 
than the STL model, 
and most often has the best performance
for the corresponding adversarial setting.
Since a shallow keypoint head
has relatively lower learning capacity,
a higher $\lambda_K$ value
will force the shared backbone
to focus excessively on the keypoint detection task,
causing deterioration in the robust MTL features
learned for the tracking task.
From \Cref{tab:carla_patch},
although we observe better generalization
for the MTL model with $\lambda_K=1.0$ 
in the benign case (mIoU = 72.08),
the adversarial robustness quickly gives away
(at 100 steps for $\delta=0.2$).
Conversely,
a lower value of $\lambda_K=0.2$
offers the best trade-off
for generalization and robustness.
\subsection{Increasing Depth of Keypoint Head}
\label{sec:res_keypoint_head_depth}

\begin{table}[t]
\centering
\caption{Ablation study with the ARMORY-CARLA dataset for attack step size $\delta=0.1$. We report the adversarial mIoU results ($\uparrow$ is better).}
\begin{tabular}{c|c|cc|cc|cc|cc} 
\toprule
 & \ $\lambda_K = 0.0$ \ & \multicolumn{4}{c|}{$\lambda_K = 0.2$} & \multicolumn{4}{c}{$\lambda_K = 1.0$} \\
 & & \multicolumn{2}{c|}{\ not pre-trained \ } & \multicolumn{2}{c|}{pre-trained} & \multicolumn{2}{c|}{\ not pre-trained \ } & \multicolumn{2}{c}{pre-trained} \\
Steps \ & (STL) & \ shallow & \ deep \ & \ shallow & \ deep \ & \ shallow & \ deep \ & \ shallow & \ deep \ \\ 
\midrule
0 & 69.45 & {\cellcolor{mtlorange}}{69.59} & {\color{gray} 66.85} & {\cellcolor{mtlorange}}{69.62} & {\color{gray} 69.36} & {\cellcolor{mtlorange}}{\textbf{72.08}} & {\color{gray} 67.28} & {\color{gray} 64.14} & {\color{gray} 69.40} \\
\midrule
10 & 48.25 & {\cellcolor{mtlorange}}{51.88} & {\color{gray} 45.28} & {\color{gray} 47.05} & {\cellcolor{mtlorange}}{48.70} & {\cellcolor{mtlorange}}{49.50} & {\cellcolor{mtlorange}}{\textbf{55.46}} & {\cellcolor{mtlorange}}{49.91} & {\cellcolor{mtlorange}}{49.77} \\
20 & 40.70 & {\cellcolor{mtlorange}}{41.44} & {\color{gray} 38.54} & {\color{gray} 39.94} & {\color{gray} 40.47} & {\cellcolor{mtlorange}}{44.47} & {\cellcolor{mtlorange}}{\textbf{47.44}} & {\cellcolor{mtlorange}}{43.10} & {\cellcolor{mtlorange}}{42.55} \\
50 & 32.07 & {\cellcolor{mtlorange}}{33.04} & {\color{gray} 31.71} & {\cellcolor{mtlorange}}{34.11} & {\cellcolor{mtlorange}}{35.28} & {\cellcolor{mtlorange}}{34.49} & {\cellcolor{mtlorange}}{\textbf{36.52}} & {\cellcolor{mtlorange}}{36.31} & {\cellcolor{mtlorange}}{32.96} \\
100 & 26.57 & {\cellcolor{mtlorange}}{28.16} & {\cellcolor{mtlorange}}{27.21} & {\cellcolor{mtlorange}}{31.33} & {\cellcolor{mtlorange}}{32.04} & {\cellcolor{mtlorange}}{30.62} & {\cellcolor{mtlorange}}{30.07} & {\cellcolor{mtlorange}}{\textbf{32.14}} & {\cellcolor{mtlorange}}{31.60} \\
200 & 24.72 & {\cellcolor{mtlorange}}{25.19} & {\color{gray} 24.15} & {\cellcolor{mtlorange}}{25.67} & {\cellcolor{mtlorange}}{25.34} & {\color{gray} 22.12} & {\cellcolor{mtlorange}}{\textbf{26.53}} & {\cellcolor{mtlorange}}{25.16} & {\cellcolor{mtlorange}}{24.77} \\
\bottomrule
\multicolumn{10}{l}{\scriptsize \hlo{orange} = MTL $>$ STL; {\color{gray} gray} = MTL $\leq$ STL; \textbf{bold} = highest in row}
\end{tabular}
\label{tab:ablation}
\end{table}

Following the observations 
with a shallow keypoint head architecture,
we also experiment with increasing
the depth of the keypoint head
from $\{128, 64\}$ channels
to $\{128, 128, 64, 64\}$ channels,
doubling the parameters
of the keypoint head network.
\Cref{tab:ablation} shows the ablation
results for the shallow and deep keypoint heads
with the ARMORY-CARLA dataset for
$\lambda_K=\{0.2, 1.0\}$
and $\delta=0.1$.
In this section
we will focus on the
``not pre-trained"
results.
The robustness of the MTL model
degrades for $\lambda_K=0.2$
when the model depth is increased,
and is easier to attack compared
to the STL model.
However, the MTL model
with deeper keypoint head has
the best adversarial robustness
for a higher $\lambda_K=1.0$,
even outperforming
the MTL model with shallow keypoint head 
for $\lambda_K=0.2$.
As the deeper keypoint head
has a relatively higher learning capacity,
it can learn to detect keypoints
with smaller changes to the 
feature space of the backbone network.
Hence, a higher $\lambda_K$
is required to adequately
supervise the backbone
in learning robust MTL features.
Although we see a decline in the
benign mIoU for increasing
depth, the deep MTL model
with $\lambda_K=1.0$
has overall best robustness.
On the other hand,
the shallow MTL model
with $\lambda_K=0.2$
has better adversarial robustness 
than the STL model
as well as better benign performance.
\subsection{Pre-training the Keypoint Head}
\label{sec:res_keypoint_head_pretrain}

As we start with a pre-trained SiamRPN model
and an untrained keypoint head,
we also study the impact of pre-training the keypoint head
before performing MTL fine-tuning.
\Cref{tab:ablation} shows the
results of this ablation study.
We report the MTL performance
with and without pre-training
the keypoint head
with the ARMORY-CARLA dataset for
$\lambda_K=\{0.2, 1.0\}$
and $\delta=0.1$.
For the shallow keypoint head architecture,
we see minor improvements
in the MTL performance
for a higher value of $\lambda_K=1.0$,
especially at higher number of attack steps.
However, there is a sharp decrease
in the benign performance (benign mIoU = 64.14).
On the other hand,
the deep keypoint head architecture
shows relative improvement
with pre-training
for a lower value of $\lambda_K=0.2$
(benign mIoU = 69.36).
Overall, there is no significant advantage
observed from pre-training the keypoint head.
A pre-trained keypoint head
would have lower potential
to significantly modify
the learned feature space of the shared backbone
as it is already near the optima for the keypoint loss space.

\section{Conclusion}

We perform an extensive set of experiments
with adversarial attacks for 
the task of person tracking
to study the impact of multi-task learning.
Our experiments 
on simulated and real-world datasets
reveal that models 
trained with multi-task learning
for the semantically analogous
tasks of person tracking and human keypoint detection
are more resilient to 
physically realizable adversarial attacks.
Our work is the first to uncover
the robustness gains
from multi-task learning
in the context of person tracking
for physically realizable attacks.
As the tracking use case has widely ranging
real-world applications,
the threat of adversarial attacks
has equally severe implications.
We hope our work triggers new research
in this direction to further secure tracking models
from adversarial attacks.

\clearpage

\bibliographystyle{splncs04}
\bibliography{references}

\begin{thebibliography}{10}
\providecommand{\url}[1]{\texttt{#1}}
\providecommand{\urlprefix}{URL }
\providecommand{\doi}[1]{https://doi.org/#1}

\bibitem{ahmed2021real}
Ahmed, I., Jeon, G.: A real-time person tracking system based on siammask
  network for intelligent video surveillance. Journal of Real-Time Image
  Processing  \textbf{18}(5),  1803--1814 (2021)

\bibitem{bertinetto2016fully}
Bertinetto, L., Valmadre, J., Henriques, J.F., Vedaldi, A., Torr, P.H.:
  Fully-convolutional siamese networks for object tracking. In: European
  conference on computer vision. pp. 850--865. Springer (2016)

\bibitem{bhattacharyya2018long}
Bhattacharyya, A., Fritz, M., Schiele, B.: Long-term on-board prediction of
  people in traffic scenes under uncertainty. In: Proceedings of the IEEE
  Conference on Computer Vision and Pattern Recognition. pp. 4194--4202 (2018)

\bibitem{bohush2019robust}
Bohush, R., Zakharava, I.: Robust person tracking algorithm based on
  convolutional neural network for indoor video surveillance systems. In:
  International Conference on Pattern Recognition and Information Processing.
  pp. 289--300. Springer (2019)

\bibitem{bridgeman2019multi}
Bridgeman, L., Volino, M., Guillemaut, J.Y., Hilton, A.: Multi-person 3d pose
  estimation and tracking in sports. In: Proceedings of the IEEE/CVF conference
  on computer vision and pattern recognition workshops. pp.~0--0 (2019)

\bibitem{chen2018shapeshifter}
Chen, S.T., Cornelius, C., Martin, J., Chau, D.H.P.: Shapeshifter: Robust
  physical adversarial attack on {F}aster {R}-{CNN} object detector. In: Joint
  European Conference on Machine Learning and Knowledge Discovery in Databases.
  pp. 52--68. Springer (2018)

\bibitem{chen2021unified}
Chen, X., Fu, C., Zheng, F., Zhao, Y., Li, H., Luo, P., Qi, G.J.: A unified
  multi-scenario attacking network for visual object tracking. In: Proceedings
  of the AAAI Conference on Artificial Intelligence. vol.~35, pp. 1097--1104
  (2021)

\bibitem{cisse2017parseval}
Cisse, M., Bojanowski, P., Grave, E., Dauphin, Y., Usunier, N.: Parseval
  networks: Improving robustness to adversarial examples. In: International
  Conference on Machine Learning. pp. 854--863. PMLR (2017)

\bibitem{dong2018boosting}
Dong, Y., Liao, F., Pang, T., Su, H., Zhu, J., Hu, X., Li, J.: Boosting
  adversarial attacks with momentum. In: Proceedings of the IEEE conference on
  computer vision and pattern recognition. pp. 9185--9193 (2018)

\bibitem{dosovitskiy17carla}
Dosovitskiy, A., Ros, G., Codevilla, F., Lopez, A., Koltun, V.: {CARLA}: {An}
  open urban driving simulator. In: Proceedings of the 1st Annual Conference on
  Robot Learning. pp. 1--16 (2017)

\bibitem{eykholt2018robust}
Eykholt, K., Evtimov, I., Fernandes, E., Li, B., Rahmati, A., Xiao, C.,
  Prakash, A., Kohno, T., Song, D.: Robust physical-world attacks on deep
  learning visual classification. In: Proceedings of the IEEE conference on
  computer vision and pattern recognition. pp. 1625--1634 (2018)

\bibitem{fan2019lasot}
Fan, H., Lin, L., Yang, F., Chu, P., Deng, G., Yu, S., Bai, H., Xu, Y., Liao,
  C., Ling, H.: La{SOT}: A high-quality benchmark for large-scale single object
  tracking. In: Proceedings of the IEEE/CVF conference on computer vision and
  pattern recognition. pp. 5374--5383 (2019)

\bibitem{ghamizi2021adversarial}
Ghamizi, S., Cordy, M., Papadakis, M., Traon, Y.L.: Adversarial robustness in
  multi-task learning: Promises and illusions. arXiv preprint arXiv:2110.15053
  (2021)

\bibitem{goodfellow2014explaining}
Goodfellow, I.J., Shlens, J., Szegedy, C.: Explaining and harnessing
  adversarial examples. arXiv preprint arXiv:1412.6572  (2014)

\bibitem{guo2017learning}
Guo, Q., Feng, W., Zhou, C., Huang, R., Wan, L., Wang, S.: Learning dynamic
  siamese network for visual object tracking. In: Proceedings of the IEEE
  international conference on computer vision. pp. 1763--1771 (2017)

\bibitem{held2016learning}
Held, D., Thrun, S., Savarese, S.: Learning to track at 100 fps with deep
  regression networks. In: European conference on computer vision. pp.
  749--765. Springer (2016)

\bibitem{ilyas2019adversarial}
Ilyas, A., Santurkar, S., Tsipras, D., Engstrom, L., Tran, B., Madry, A.:
  Adversarial examples are not bugs, they are features. Advances in neural
  information processing systems  \textbf{32} (2019)

\bibitem{jia2020robust}
Jia, S., Ma, C., Song, Y., Yang, X.: Robust tracking against adversarial
  attacks. In: European Conference on Computer Vision. pp. 69--84. Springer
  (2020)

\bibitem{jia2021iou}
Jia, S., Song, Y., Ma, C., Yang, X.: Iou attack: Towards temporally coherent
  black-box adversarial attack for visual object tracking. In: Proceedings of
  the IEEE/CVF Conference on Computer Vision and Pattern Recognition. pp.
  6709--6718 (2021)

\bibitem{jia2020fooling}
Jia, Y.J., Lu, Y., Shen, J., Chen, Q.A., Chen, H., Zhong, Z., Wei, T.W.:
  Fooling detection alone is not enough: Adversarial attack against multiple
  object tracking. In: International Conference on Learning Representations
  (ICLR'20) (2020)

\bibitem{kocabas2018multiposenet}
Kocabas, M., Karagoz, S., Akbas, E.: Multiposenet: Fast multi-person pose
  estimation using pose residual network. In: Proceedings of the European
  conference on computer vision (ECCV). pp. 417--433 (2018)

\bibitem{kong2020long}
Kong, L., Huang, D., Wang, Y.: Long-term action dependence-based hierarchical
  deep association for multi-athlete tracking in sports videos. IEEE
  Transactions on Image Processing  \textbf{29},  7957--7969 (2020)

\bibitem{kristan2020eighth}
Kristan, M., Leonardis, A., Matas, J., Felsberg, M., Pflugfelder, R.,
  K{\"a}m{\"a}r{\"a}inen, J.K., Danelljan, M., Zajc, L.{\v{C}}.,
  Luke{\v{z}}i{\v{c}}, A., Drbohlav, O., et~al.: The eighth visual object
  tracking vot2020 challenge results. In: European Conference on Computer
  Vision. pp. 547--601. Springer (2020)

\bibitem{kristan2021ninth}
Kristan, M., Matas, J., Leonardis, A., Felsberg, M., Pflugfelder, R.,
  K{\"a}m{\"a}r{\"a}inen, J.K., Chang, H.J., Danelljan, M., Cehovin, L.,
  Luke{\v{z}}i{\v{c}}, A., et~al.: The ninth visual object tracking vot2021
  challenge results. In: Proceedings of the IEEE/CVF International Conference
  on Computer Vision. pp. 2711--2738 (2021)

\bibitem{kristan2019seventh}
Kristan, M., Matas, J., Leonardis, A., Felsberg, M., Pflugfelder, R.,
  Kamarainen, J.K., ˇCehovin~Zajc, L., Drbohlav, O., Lukezic, A., Berg, A.,
  et~al.: The seventh visual object tracking vot2019 challenge results. In:
  Proceedings of the IEEE/CVF International Conference on Computer Vision
  Workshops. pp.~0--0 (2019)

\bibitem{krizhevsky2012imagenet}
Krizhevsky, A., Sutskever, I., Hinton, G.E.: Imagenet classification with deep
  convolutional neural networks. Advances in neural information processing
  systems  \textbf{25} (2012)

\bibitem{li2019siamrpn++}
Li, B., Wu, W., Wang, Q., Zhang, F., Xing, J., Yan, J.: Siamrpn++: Evolution of
  siamese visual tracking with very deep networks. In: Proceedings of the
  IEEE/CVF Conference on Computer Vision and Pattern Recognition. pp.
  4282--4291 (2019)

\bibitem{li2018high}
Li, B., Yan, J., Wu, W., Zhu, Z., Hu, X.: High performance visual tracking with
  siamese region proposal network. In: Proceedings of the IEEE conference on
  computer vision and pattern recognition. pp. 8971--8980 (2018)

\bibitem{liang2020multi}
Liang, Q., Wu, W., Yang, Y., Zhang, R., Peng, Y., Xu, M.: Multi-player tracking
  for multi-view sports videos with improved k-shortest path algorithm. Applied
  Sciences  \textbf{10}(3), ~864 (2020)

\bibitem{liang2020efficient}
Liang, S., Wei, X., Yao, S., Cao, X.: Efficient adversarial attacks for visual
  object tracking. In: European Conference on Computer Vision. pp. 34--50.
  Springer (2020)

\bibitem{liebel2019multidepth}
Liebel, L., K{\"o}rner, M.: Multidepth: Single-image depth estimation via
  multi-task regression and classification. In: 2019 IEEE Intelligent
  Transportation Systems Conference (ITSC). pp. 1440--1447. IEEE (2019)

\bibitem{lin2014coco}
Lin, T.Y., Maire, M., Belongie, S., Hays, J., Perona, P., Ramanan, D.,
  Doll{\'a}r, P., Zitnick, C.L.: Microsoft coco: Common objects in context. In:
  Fleet, D., Pajdla, T., Schiele, B., Tuytelaars, T. (eds.) Computer Vision --
  ECCV 2014. pp. 740--755. Springer International Publishing, Cham (2014)

\bibitem{liu2017adversarial}
Liu, P., Qiu, X., Huang, X.: Adversarial multi-task learning for text
  classification. arXiv preprint arXiv:1704.05742  (2017)

\bibitem{luo2013manifold}
Luo, Y., Tao, D., Geng, B., Xu, C., Maybank, S.J.: Manifold regularized
  multitask learning for semi-supervised multilabel image classification. IEEE
  Transactions on Image Processing  \textbf{22}(2),  523--536 (2013)

\bibitem{madry2017towards}
Madry, A., Makelov, A., Schmidt, L., Tsipras, D., Vladu, A.: Towards deep
  learning models resistant to adversarial attacks. arXiv preprint
  arXiv:1706.06083  (2017)

\bibitem{mao2020multitask}
Mao, C., Gupta, A., Nitin, V., Ray, B., Song, S., Yang, J., Vondrick, C.:
  Multitask learning strengthens adversarial robustness. In: Computer Vision -
  {ECCV} 2020 - 16th European Conference, Glasgow, UK, August 23-28, 2020,
  Proceedings, Part {II}. Lecture Notes in Computer Science, vol. 12347, pp.
  158--174. Springer (2020)

\bibitem{art2018}
Nicolae, M.I., Sinn, M., Tran, M.N., Buesser, B., Rawat, A., Wistuba, M.,
  Zantedeschi, V., Baracaldo, N., Chen, B., Ludwig, H., Molloy, I., Edwards,
  B.: Adversarial robustness toolbox v1.2.0. CoRR  \textbf{1807.01069} (2018),
  \url{https://arxiv.org/pdf/1807.01069}

\bibitem{ren2015faster}
Ren, S., He, K., Girshick, R., Sun, J.: Faster {R}-{CNN}: Towards real-time
  object detection with region proposal networks. Advances in neural
  information processing systems  \textbf{28} (2015)

\bibitem{schmidt2018adversarially}
Schmidt, L., Santurkar, S., Tsipras, D., Talwar, K., Madry, A.: Adversarially
  robust generalization requires more data. Advances in neural information
  processing systems  \textbf{31} (2018)

\bibitem{shuai2021siammot}
Shuai, B., Berneshawi, A., Li, X., Modolo, D., Tighe, J.: Siammot: Siamese
  multi-object tracking. In: Proceedings of the IEEE/CVF conference on computer
  vision and pattern recognition. pp. 12372--12382 (2021)

\bibitem{simon2019first}
Simon-Gabriel, C.J., Ollivier, Y., Bottou, L., Sch{\"o}lkopf, B., Lopez-Paz,
  D.: First-order adversarial vulnerability of neural networks and input
  dimension. In: International Conference on Machine Learning. pp. 5809--5817.
  PMLR (2019)

\bibitem{szegedy2013intriguing}
Szegedy, C., Zaremba, W., Sutskever, I., Bruna, J., Erhan, D., Goodfellow, I.,
  Fergus, R.: Intriguing properties of neural networks. arXiv preprint
  arXiv:1312.6199  (2013)

\bibitem{tramer2017ensemble}
Tram{\`e}r, F., Kurakin, A., Papernot, N., Goodfellow, I., Boneh, D., McDaniel,
  P.: Ensemble adversarial training: Attacks and defenses. arXiv preprint
  arXiv:1705.07204  (2017)

\bibitem{armory2022}
{Two Six Technologies}: {ARMORY}. \url{https://github.com/twosixlabs/armory}

\bibitem{wang2021towards}
Wang, X., Shu, X., Zhang, Z., Jiang, B., Wang, Y., Tian, Y., Wu, F.: Towards
  more flexible and accurate object tracking with natural language: Algorithms
  and benchmark. In: Proceedings of the IEEE/CVF Conference on Computer Vision
  and Pattern Recognition. pp. 13763--13773 (2021)

\bibitem{wiyatno2019physical}
Wiyatno, R.R., Xu, A.: Physical adversarial textures that fool visual object
  tracking. In: Proceedings of the IEEE/CVF International Conference on
  Computer Vision. pp. 4822--4831 (2019)

\bibitem{wu2015otb}
Wu, Y., Lim, J., Yang, M.H.: Object tracking benchmark. IEEE Transactions on
  Pattern Analysis and Machine Intelligence  \textbf{37}(9),  1834--1848
  (2015). \doi{10.1109/TPAMI.2014.2388226}

\bibitem{xu2020adversarial_attack}
Xu, H., Ma, Y., Liu, H.C., Deb, D., Liu, H., Tang, J.L., Jain, A.K.:
  Adversarial attacks and defenses in images, graphs and text: A review.
  International Journal of Automation and Computing  \textbf{17}(2),  151--178
  (2020)

\bibitem{xu2020adversarial}
Xu, K., Zhang, G., Liu, S., Fan, Q., Sun, M., Chen, H., Chen, P.Y., Wang, Y.,
  Lin, X.: Adversarial {T}-shirt! evading person detectors in a physical world.
  In: European conference on computer vision. pp. 665--681. Springer (2020)

\bibitem{yagi2018future}
Yagi, T., Mangalam, K., Yonetani, R., Sato, Y.: Future person localization in
  first-person videos. In: Proceedings of the IEEE Conference on Computer
  Vision and Pattern Recognition. pp. 7593--7602 (2018)

\bibitem{yan2021depthtrack}
Yan, S., Yang, J., K{\"a}pyl{\"a}, J., Zheng, F., Leonardis, A.,
  K{\"a}m{\"a}r{\"a}inen, J.K.: Depthtrack: Unveiling the power of rgbd
  tracking. In: Proceedings of the IEEE/CVF International Conference on
  Computer Vision. pp. 10725--10733 (2021)

\bibitem{yan2018deep}
Yan, Z., Guo, Y., Zhang, C.: Deep defense: Training dnns with improved
  adversarial robustness. Advances in Neural Information Processing Systems
  \textbf{31} (2018)

\bibitem{yao2019unsupervised}
Yao, Y., Xu, M., Wang, Y., Crandall, D.J., Atkins, E.M.: Unsupervised traffic
  accident detection in first-person videos. In: 2019 IEEE/RSJ International
  Conference on Intelligent Robots and Systems (IROS). pp. 273--280. IEEE
  (2019)

\bibitem{ye2020person}
Ye, S., Bohush, R., Chen, H., Zakharava, I.Y., Ablameyko, S.: Person tracking
  and reidentification for multicamera indoor video surveillance systems.
  Pattern Recognition and Image Analysis  \textbf{30}(4),  827--837 (2020)

\bibitem{yu2020deformable}
Yu, Y., Xiong, Y., Huang, W., Scott, M.R.: Deformable siamese attention
  networks for visual object tracking. In: Proceedings of the IEEE/CVF
  Conference on Computer Vision and Pattern Recognition. pp. 6728--6737 (2020)

\bibitem{zhang2013robust}
Zhang, T., Ghanem, B., Liu, S., Ahuja, N.: Robust visual tracking via
  structured multi-task sparse learning. International journal of computer
  vision  \textbf{101}(2),  367--383 (2013)

\bibitem{zhang2017multi}
Zhang, T., Xu, C., Yang, M.H.: Multi-task correlation particle filter for
  robust object tracking. In: Proceedings of the IEEE conference on computer
  vision and pattern recognition. pp. 4335--4343 (2017)

\bibitem{zhang2018learning}
Zhang, T., Xu, C., Yang, M.H.: Learning multi-task correlation particle filters
  for visual tracking. IEEE transactions on pattern analysis and machine
  intelligence  \textbf{41}(2),  365--378 (2018)

\bibitem{zhang2019siamft}
Zhang, X., Ye, P., Peng, S., Liu, J., Gong, K., Xiao, G.: Siamft: An
  rgb-infrared fusion tracking method via fully convolutional siamese networks.
  IEEE Access  \textbf{7},  122122--122133 (2019)

\bibitem{zhang2020dsiammft}
Zhang, X., Ye, P., Peng, S., Liu, J., Xiao, G.: Dsiammft: An rgb-t fusion
  tracking method via dynamic siamese networks using multi-layer feature
  fusion. Signal Processing: Image Communication  \textbf{84},  115756 (2020)

\bibitem{zhang2021survey}
Zhang, Y., Yang, Q.: A survey on multi-task learning. IEEE Transactions on
  Knowledge and Data Engineering  (2021)

\bibitem{zhu2022visual}
Zhu, X.F., Xu, T., Wu, X.J.: Visual object tracking on multi-modal rgb-d
  videos: A review. arXiv preprint arXiv:2201.09207  (2022)

\bibitem{zhu2018distractor}
Zhu, Z., Wang, Q., Li, B., Wu, W., Yan, J., Hu, W.: Distractor-aware siamese
  networks for visual object tracking. In: Proceedings of the European
  conference on computer vision (ECCV). pp. 101--117 (2018)

\end{thebibliography}
\end{document}